\documentclass[letterpaper, 10 pt, conference]{ieeeconf}  

\IEEEoverridecommandlockouts                              

\usepackage{graphics}
\usepackage{amsmath}
\usepackage{graphicx}
\usepackage{booktabs}
\usepackage{multirow}
\usepackage{hyperref}
\usepackage{tcolorbox}
\usepackage{amssymb}
\usepackage{ragged2e}
\usepackage{bbm}
\usepackage{url}
\usepackage{marvosym}
\usepackage{makecell}
\usepackage{adjustbox}
\usepackage{lipsum}

\title{\LARGE \bf
\emph{\textbf{FastUMI‑100K}}: Advancing Data-driven Robotic Manipulation \\ with a Large-scale UMI-style Dataset}
\author{\textbf{Kehui Liu$^{1,2,*}$}, \textbf{Zhongjie Jia$^{1,3,*}$},  \textbf{Yang Li$^{1,4*}$}, \textbf{Zhaxizhuoma$^{1,3*}$}\\ Pengan Chen$^{1}$, Song Liu$^{1}$, Xin Liu$^{1}$, Pingrui Zhang$^{1}$, Haoming Song$^{1}$, Xinyi Ye$^{1}$, Nieqing Cao$^{5}$\\
Zhigang Wang$^{1}$, Jia Zeng$^{1}$, Dong Wang$^{1}$, Yan Ding$^{1,6}$, Bin Zhao$^{1,2}$, Xuelong Li$^{7}$ 
\thanks{$^{1}$Shanghai Artificial Intelligence Laboratory. $^{2}$Northwestern Polytechnical University. $^{3}$Shanghai Jiao Tong University. $^{4}$TongJi University. $^{5}$Xi'an Jiaotong-Liverpool University. $^{6}$Suzhou OneStar Robotics Corp Ltd. $^{7}$Institute of Artificial Intelligence, China Telecom Corp Ltd. $*$Equal contribution.}%
\thanks{Email: \href{mailto:liukehui@pjlab.org.cn}{liukehui@pjlab.org.cn} and \href{mailto:zhaobin@pjlab.org.cn}{zhaobin@pjlab.org.cn}.}%
}

\begin{document}

\maketitle
\thispagestyle{empty}
\pagestyle{empty}

\begin{abstract}

Data-driven robotic manipulation learning depends on large-scale, high-quality expert demonstration datasets. However, existing datasets, which primarily rely on human teleoperated robot collection, are limited in terms of scalability, trajectory smoothness, and applicability across different robotic embodiments in real-world environments. In this paper, we present FastUMI-100K, a large-scale UMI-style multimodal demonstration dataset, designed to overcome these limitations and meet the growing complexity of real-world manipulation tasks. Collected by FastUMI, a novel robotic system featuring a modular, hardware-decoupled mechanical design and an integrated lightweight tracking system, FastUMI-100K offers a more scalable, flexible, and adaptable solution to fulfill the diverse requirements of real-world robot demonstration data. Specifically, FastUMI‑100K contains over 100K+ demonstration trajectories collected across representative household environments, covering 54 tasks and hundreds of object types. Our dataset integrates multimodal streams, including end-effector states, multi-view wrist-mounted fisheye images and textual annotations. Each trajectory has a length ranging from 120 to 500 frames. Experimental results demonstrate that FastUMI-100K enables high policy success rates across various baseline algorithms, confirming its robustness, adaptability, and real-world applicability for solving complex, dynamic manipulation challenges. The source code and dataset will be released in this link~\url{https://github.com/MrKeee/FastUMI-100K}.

\end{abstract}

\section{Introduction}
\begin{figure*}[t]
    \centering
    \includegraphics[width=1.0\textwidth]{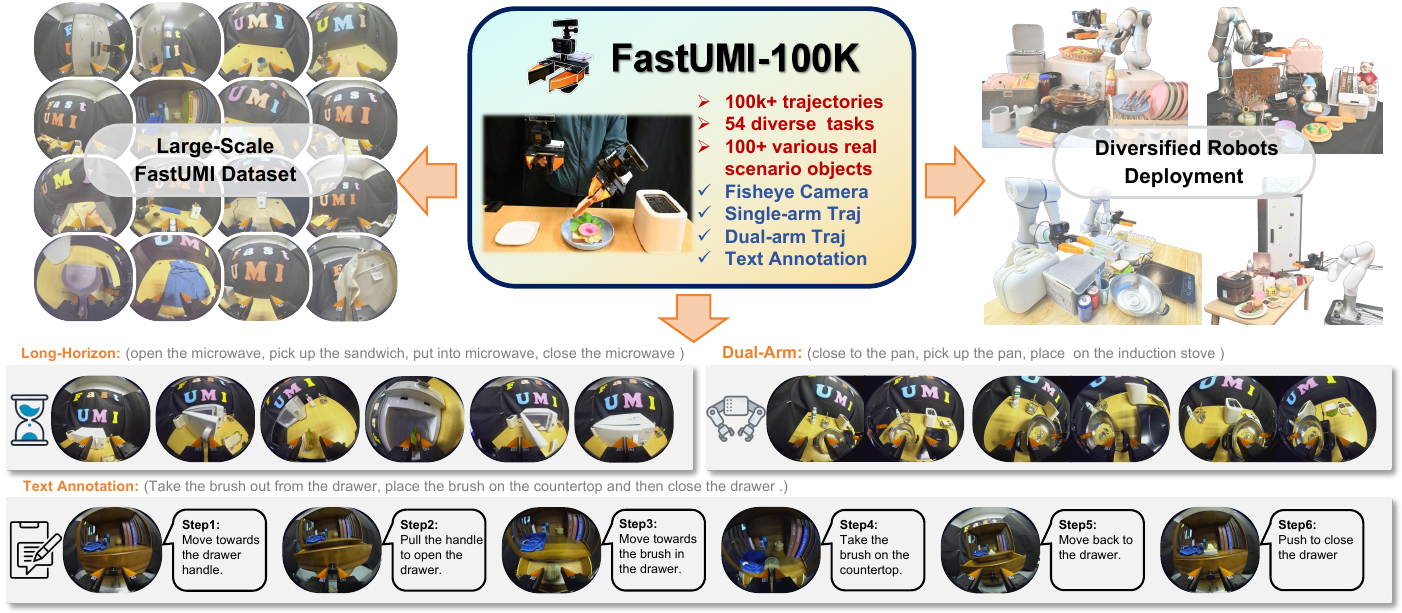}
    \vskip -0.0in  
    \caption{\textbf{Overview of FastUMI-100K.} We introduce FastUMI-100K, a large-scale UMI-style dataset. This dataset comprises 100K+ long-horizon trajectories, spanning 54 distinct tasks and over 100 real scenario manipulation objects. FastUMI-100K integrates multimodal data streams, including single-arm and dual-arm trajectories, multi-view wrist-mounted fisheye image, and fine-grained textual annotations. Benefitting from our embodiment-agnostic design, FastUMI-100K can be used across diverse robotics. }
    \label{fig:main}
    \vskip -0.25in  
\end{figure*}

Robotic Manipulation is the core capabilities of embodied intelligence. Data-driven imitation learning plays a crucial role in achieving generalized manipulation policy~\cite{black2024pi0visionlanguageactionflowmodel,qu2025embodiedonevision,qu2025spatialvla,team2024octo,liu2024rdt,kim2024openvla,song2025hume}. However, the collection of manipulation data is a time-consuming and labor-intensive process, posing a significant bottleneck in scaling up imitation learning. Therefore, constructing a large-scale, high-quality manipulation dataset for various types of robots has become an urgent need for robotics research~\cite{walke2023bridgedata,o2024open,wu2024robomind,bu2025agibot,fang2023rh20t,ebert2021bridge}.

Although the collection of robot data faces challenges such as high costs of hardware and human labor, several existing works have made significant efforts to curate datasets for robots, aiming to advance the training of robotic policy models~\cite{walke2023bridgedata,o2024open,brohan2022rt}. However, despite the growing number of existing datasets, the amount of data available for robots remains relatively small when compared with the vast datasets used for training general models in fields such as computer vision and natural language processing. Moreover, the current available robot data collected with laborious teleoperation focuses on relatively simple tasks, such as pick-and-place operations, significant challenges remain in the collection of more complex, fine-grained, dual-arm collaborative, and long-horizon tasks, which requires more efficient and scalable data collection methods.

Current data collection methods can be broadly categorized into teleoperation-based techniques~\cite{wu2024gello,iyer2024open}, vision-driven demonstration methods~\cite{morgan2021vision,wen2022you}, and sensor-enhanced interfaces~\cite{song2020grasping,chi2024universal}. The teleoperation-based approach, which requires full-process manual remote control, not only increases the operator's labor intensity but also significantly raises both labor and time costs. The vision-driven demonstration method, while useful, lacks robot action labels and fails to capture fine-grained dynamic data (\emph{e.g.}, force feedback, contact states) during robot-object interactions. In contrast, the sensor-enhanced interface converts human demonstrations into robot executable data with high authenticity and accuracy, effectively addressing the adaptation challenge between human operations and robot replication. Specifically, building on UMI~\cite{chi2024universal}, FastUMI~\cite{liu2024fastumi} replaces the GoPro-based Visual Inertial Odometry (VIO) pipeline and open-source SLAM algorithms with the existing Realsense T265 camera, thus enabling faster collection of high-quality data under real-world scenarios. In this way, the FastUMI data collection system bridges the gap between human demonstration and autonomous robot execution, facilitating rapid and high-quality data collection.

Utilizing the FastUMI data collection system~\cite{liu2024fastumi}, we integrated single-arm and dual-arm configurations with adaptable universal finger sleeves to conduct large-scale data collection. In this paper, we introduce the large-scale UMI-style multimodal dataset---FastUMI-100K, which incorporates the dataset of the pioneering work FastUMI and totally comprises over 100,000 demonstration trajectories, collected using both single-arm and dual-arm grippers on the FastUMI platform, equivalent to 600 hours of interactive data. A total of 10 data collectors and 3 technical support staff were employed, and 5 standardized collection environments were established. Data collection was conducted simultaneously by five teams, each consisting of two members responsible for data collection, code program switching, data visualization, and real-time monitoring of data quality. The FastUMI-100K dataset encompasses 54 distinct task scenarios, covering both single-arm and dual-arm tasks across a range of common household environments and involving hundreds of different real scenario objects. Each trajectory consists of 120 to 500 video frames and includes multimodal data across various modalities including single-arm and dual-arm robot state information, multi-view fisheye images and textual annotations. The dataset contains tasks such as multi-object pick-and-place, hinged object handling, dual-arm gripper collaboration, and fine object manipulation, involving long-horizon tasks, aiming to replicate challenging real-world scenarios. Fine-grained temporal alignment across multiple sensors has been achieved to decouple sensor data. A comprehensive data processing tool chain is provided, adaptable to various robot policy learning algorithms.
\section{Related Work}
\subsection{Sensor-enhanced Interface Dataset}
Handheld, interface–mediated collection offers a portable alternative to teleoperation and third-person video. UMI shows that a hand-held gripper with a wrist-mounted camera and an embodiment-agnostic policy interface can capture information-dense demonstrations for dynamic, long-horizon, and bimanual skills while matching the deployment viewpoint~\cite{chi2024universal}. Building on this paradigm, Fast-UMI increases throughput and cross-embodiment transfer by decoupling handheld hardware from robot end-effectors, retaining the wrist fisheye view, and directly logging 6-DoF end-effector poses with RealSense~T265—avoiding heavy offline SLAM or fixed motion-capture infrastructure~\cite{liu2024fastumi}. Beyond vision-only interfaces, portable visuo-tactile devices further enrich contact cues: ViTaMIn introduces an embodiment-free visuo-tactile gripper with pretraining for tactile representations; FreeTacMan proposes a wearable dual visuo-tactile setup with external optical tracking; and 3D-ViTac highlights joint vision–touch learning for fine, bimanual manipulation~\cite{liu2025vitamin,wu2025freetacman,huang20243d}. In contrast, recent teleoperation systems—GELLO (leader–follower), ALOHA/ALOHA~2 (low-cost bimanual puppeteering), and Bunny-VisionPro (bimanual teleop with haptics)—provide high-fidelity control but remain hardware-coupled at collection time and require device-specific remapping for reuse~\cite{wu2024gello,aldaco2024aloha,ding2024bunny}. Fast-UMI targets these pains via hardware-decoupled handheld devices and deployment-time wrist views~\cite{liu2024fastumi}.

\subsection{Large-Scale Data in Policy Learning}
Scaling data diversity consistently improves generalization and zero-/few-shot robustness. Early cross-robot corpora such as RoboNet standardized multi-robot visual interaction; BridgeData~V2 expanded multi-environment, instruction-conditioned behaviors; and DROID pushed in-the-wild diversity to 76k demonstrations across 500+ scenes~\cite{dasari2019robonet,walke2023bridgedata,khazatsky2024droid}. The Open X-Embodiment collaboration aggregated $1$M$+$ trajectories over 22 embodiments and trained RT-X models that exhibit cross-embodiment transfer~\cite{o2024open}. On the model side, RT-1/RT-2 demonstrate the benefits of large robot corpora and web-scale vision–language pretraining, while generalist VLA policies such as $\pi_0$ adopt flow-matching heads atop internet-pretrained VLMs and co-train on heterogeneous robot datasets to further enhance transfer~\cite{brohan2022rt,zitkovich2023rt,black2024pi0visionlanguageactionflowmodel}. Within this landscape, UMI-style handheld datasets offer complementary advantages. They preserve the deployment wrist viewpoint, provide smooth low-latency end-effector state without lab infrastructure, and are inherently hardware-decoupled that contribute to higher dataset utility and potential for reuse~\cite{belkhale2023data}. FastUMI-100K extends this niche with a large-scale, multimodal UMI-style corpus that includes single-arm and dual-arm configurations, and hierarchical textual annotations. The dataset spans long-horizon, hinged, and deformable-object tasks, supporting single-task imitation learning, cross-embodiment transfer through simple frame mapping, and generalist Vision-Language-Action (VLA) model fine-tuning~\cite{black2024pi0visionlanguageactionflowmodel,liu2024fastumi}.

\section{Data Collection and Processing}
\subsection{Hardware Design}
\label{subsec: hardware design}

\begin{figure}
\centering
\includegraphics[width=0.48\textwidth]{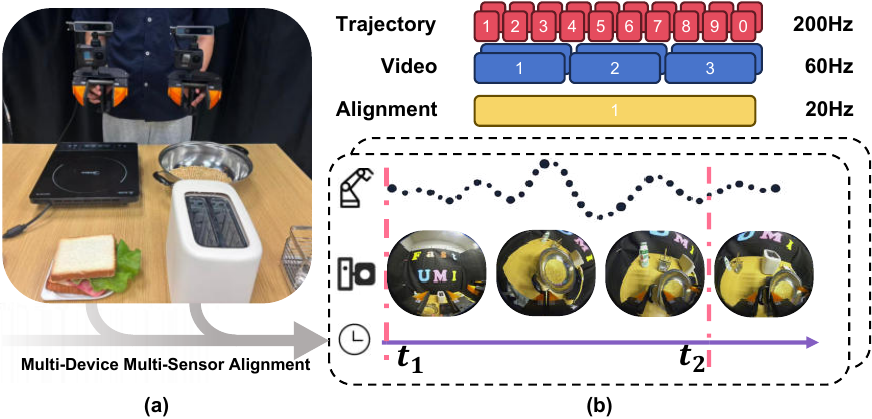}
\vskip -0.05in  
\caption{Left figure shows the dual-arm FastUMI hardware data collection device developed in our dataset. Right figure shows the schematic diagram of multi-sensor temporal alignment at 20 hz frequency.}
\label{fig:hardware}
\vskip -0.25in    
\end{figure}

Following the protocols established in FastUMI, we use the RealSense T265 for trajectory tracking, while high-resolution, wide-angle fisheye RGB images are captured using GoPro fisheye cameras. A standardized, plug-and-play fingertip attachment, designed to be compatible with common handheld device accessories, facilitates easy installation on various robotic grippers. This ensures that the collected data can be directly adapted to devices equipped with such fingertip attachments. As shown in Figure~\ref{fig:hardware}(a), a dual-arm FastUMI system is further developed. In this system, two T265 cameras and two GoPro fisheye cameras are integrated into the end effectors of both grippers to synchronously record multimodal data at a 20Hz sampling frequency.

\subsection{Data Collection Pipeline}
After undergoing specialized training, a dedicated FastUMI data collection team is established, comprising 10 data collectors and 3 technical support staff members, with ages ranging from 20 to 24 years. To ensure efficient and organized data collection, the team is divided into five subgroups, each consisting of two data collectors. These subgroups are responsible for tasks including data collection, code program switching, data visualization, and real-time monitoring of data quality. The three technical support staff members focus on equipment maintenance, task design, and overall management. Figure~\ref{fig:collection_pipeline} details the following three stages of data collection.

\subsubsection{Pre-Data Collection}
Before data collection begins, the technical support staff complete several preparatory tasks, which include task design, distribution, and feasibility validation, as well as setting a reference time for the collection based on task difficulty. To ensure consistency and standardization in the data collection process, technical support staff recorde instructional demonstration videos to clarify the collection protocol. Furthermore, experienced technical support staff carefully set up diverse task scenarios to enrich the variety of collected data.
\subsubsection{During Data Collection}
A strict process is followed during data collection. Two fixed 3D-printed slots are used to secure the handheld devices on the collection platform, which also serve as the initial positioning points for the T265 cameras upon startup. The initial slots for each scene are placed at consistent distances apart to allow accurate calculation of the relative positions of the arms along the full trajectory. To ensure the correctness of the collection process, we implement a voice-guided system that provides reminders at different stages of the collection. 

In parallel, real-time visualization of the GoPro fisheye images and T265 camera trajectories is displayed on the computer, enabling the operator to monitor potential issues such as frame drops or trajectory drift during data collection.
\subsubsection{Post-Data Collection}

To provide real-time feedback on data quality, a quality monitoring system is incorporated into the project. When the T265 camera data encounters quality issues, these are typically manifested as sudden trajectory jumps or global scale distortion. By calculating the linear and angular velocities between adjacent points on the collected trajectory, the system could identify operational anomalies that could lead to SLAM trajectory drift and the trajectory will be automatically marked as invalid and deleted. This self-checking mechanism encourages data collectors to adjust their actions promptly, ensuring high-quality data.
\subsection{Multi-Device Multi-Sensor Alignment}

\begin{figure}
\centering
\includegraphics[width=0.47\textwidth]{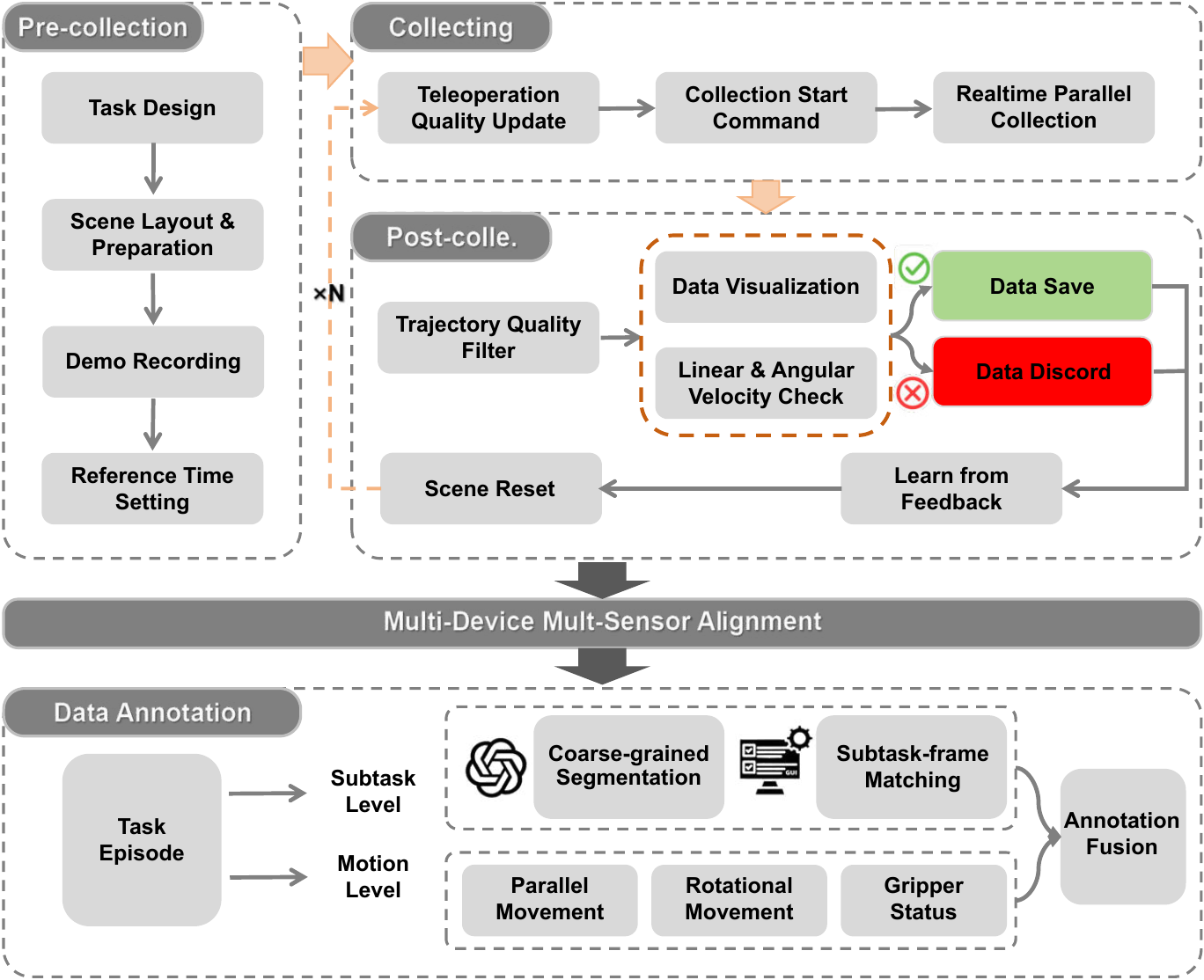}
\vskip -0.05in  
\caption{The pipeline of data collection and processing.}
\label{fig:collection_pipeline}
\vskip -0.25in    
\end{figure}

As the number of sensors increases, the challenge of achieving precise temporal alignment due to different sampling rates and data patterns in multi-device multi-sensor data fusion becomes more pronounced. To address this issue, we employ a unified Robot Operating System (ROS) clock to assign consistent timestamps across all data streams and extend the FastUMI single-arm multi-sensor alignment technique to a dual-arm scenario.
As shown in Figure~\ref{fig:hardware}(b), This system incorporates four sensors with different sampling rates, \emph{i.e.}, two GoPro cameras at 60Hz and two T265 sensors at 200Hz. The specific alignment strategy is as follows: using ROS's approximate time synchronizer, we perform approximate synchronization of the two RGB image topics for the grippers based on their timestamps. The maximum alignment error is set to half the GoPro sampling period (i.e., 1/120 seconds). Simultaneously, video frames from each gripper are uniformly downsampled to 20Hz and matched with the closest T265 pose data in time. This approach ultimately achieves sub-millisecond precision in the alignment of dual-arm multi-sensor data, providing a stable and synchronized dataset for downstream learning tasks.

\begin{figure*}[t]
    \centering
    \includegraphics[width=1.0\textwidth]{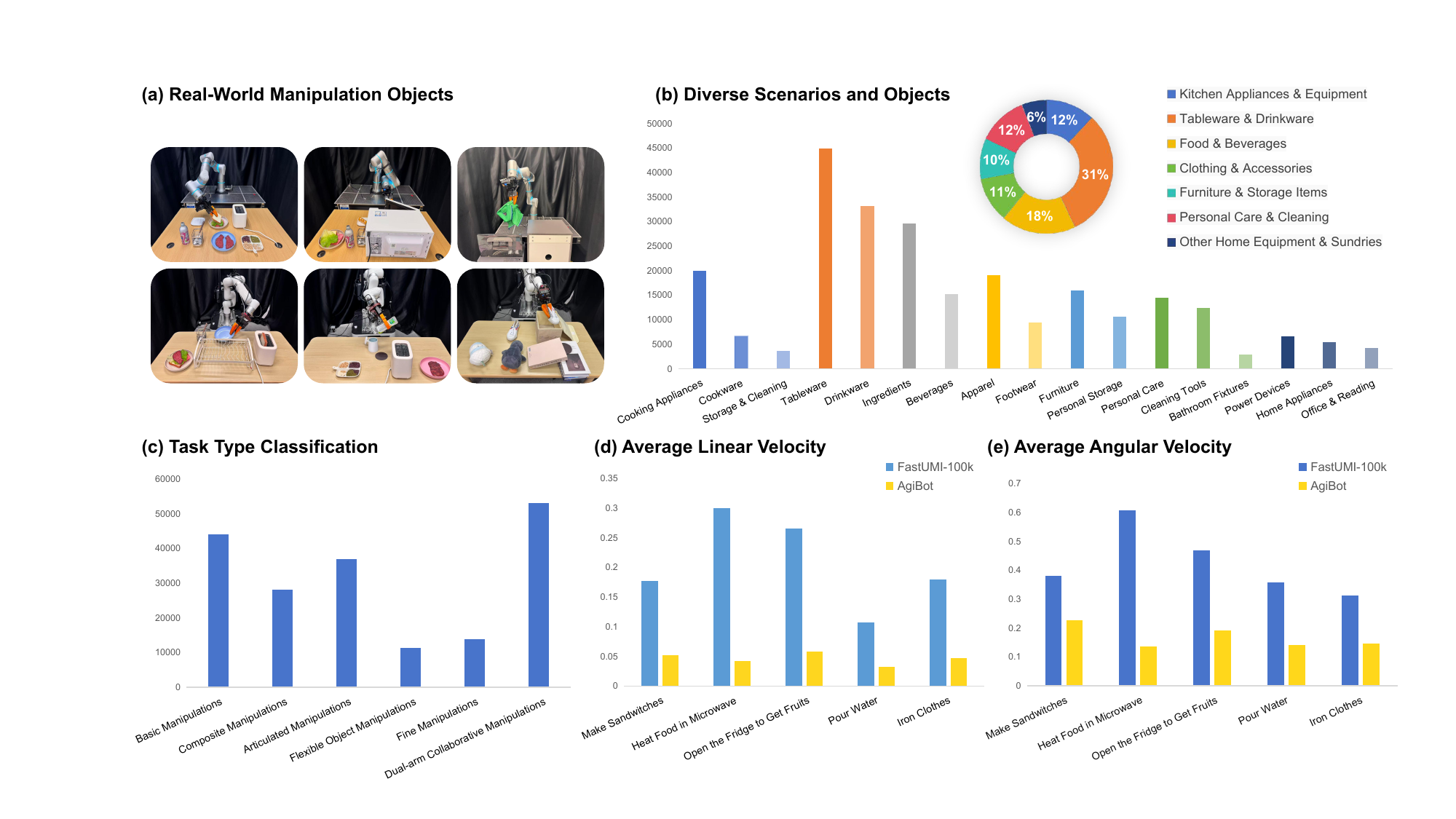}
    \vskip -0.1in  
    \caption{\textbf{Data statistical chart of FastUMI-100K.} Figure (a) shows our manipulation of objects in real-world scenarios using different robots. Figure (b) presents the classification of the number of objects across various scenarios. Figure (c) records the distribution of the number of six different types of tasks in the dataset. In Figures (d) and (e), five types of dual-arm tasks from FastUMI-100K and AgiBot are selected, and the comparison of the average linear velocity and angular velocity of their respective data is presented, demonstrating the flexibility and human-like qualities of FastUMI-100K when performing complex long-horizon tasks.}
    \label{fig:data}
    \vskip -0.2in  
\end{figure*}

\subsection{Data Annotation}
\label{subsec:data annotation}
As a critical medium for human-robot interaction, providing precise and detailed semantic information to the dataset is essential for training language-conditioned policies. To address this, we develop a duel-level text annotation system with two stages as shown in Figure~\ref{fig:collection_pipeline}: (1) breaking down long-horizon tasks into distinct phases, clarifying the manipulation objectives of each sub-phase, and (2) providing a detailed description of the robotic arm's motion-related manipulation specifics. Based on this framework, we propose two levels of annotation strategies:

\subsubsection{Subtask-Level Annotation}

Firstly, GPT-4o is utilized to analyze the corresponding video content for the complete task and automatically divides it into subtasks. After the subtask division, we design a custom GUI interface to facilitate manual annotation, extracting and segmenting key frames from trajectory images, and precisely matching each key frame with the text description of its corresponding subtask. 

Because the first-person perspective video uses a self-centered coordinate system, the subtask-level annotations primarily focus on describing the target position (\emph{e.g.}, \texttt{move towards the cup}) rather than relative position relationships (\emph{e.g.}, \texttt{move left towards the cup}) to reduce ambiguity in the annotations. The annotation process follows the mechanism of GPT-4o pre-annotation followed by manual segmentation, where multiple annotators cross-check and correct the automated annotations to ensure accuracy and consistency.
\subsubsection{Motion-Level Annotation}

Based on the RT-H~\cite{belkhale2024rt} annotation paradigm, we analyze the pose data of each frame and the tenth frame following it in three-dimensional space. We calculate the relative translation along the X, Y, and Z axes, the rotational angle changes, and the relative changes in the gripper's opening and closing. Basic semantic units such as \texttt{move forward/backward/left/right}, \texttt{rotate clockwise/counterclockwise}, and \texttt{gripper open/close} are combined to generate detailed motion descriptions, providing a precise semantic representation of the robotic arm's operation.

By adopting the duel-level text annotation system, sub-tasks and motions are annotated, enabling more detailed textual annotations and establishing a complete semantic chain from task objectives to motion details. Through cross-sample cross-validation, a total of 15,000 textual annotations are provided, ensuring the reliability of the semantic information.
\section{Dataset Analysis}

Leveraging standardized hardware and a fully structured data collection pipeline, we present FastUMI-100K, a large-scale multimodal dataset that substantially surpasses the scale and diversity of existing UMI-style datasets. FastUMI-100K is collected using both single-arm and dual-arm configurations of the FastUMI system and incorporates the dataset of the pioneering work FastUMI, comprising over 100,000 high-quality trajectory samples. It covers 54 categories of daily household tasks, hundreds of operational objects, and diverse background settings, thereby forming a comprehensive atomic skill library for general-purpose robotic manipulation. A key objective is to provide a highly diverse and generalizable dataset that is decoupled from specific robot embodiments, enabling broader applicability across platforms. 
\subsection{Diversity of Task Types}

With diverse daily household tasks, FastUMI-100K essentially forms an atomic skill library for robotic manipulation. Our task design is rooted in a core philosophy: to enable robots to truly integrate into and serve daily life, moving beyond the simple pick-and-place tasks in many existing datasets. We therefore place strong emphasis on task diversity and real-world relevance—--each task category incorporates a variety of objects, spanning hundreds of common household items such as clothing, tableware, small household appliances, cabinets, food, and sanitary fixtures. This design not only expands the dataset into a large-scale, life-oriented task corpus but also ensures the model trained on it acquires stronger generalization capability across complex, unstructured real-world scenarios. To facilitate organization and analysis, the dataset categorizes these manipulation tasks into six primary groups:

\begin{enumerate}
    \item \textbf{Basic Manipulations}: Includes simple actions such as grasping and placing single or multiple objects, without involving complex rotational movements.
    \item \textbf{Composite Manipulations}: Involves complex tasks combining parallel movement and rotation, requiring consideration of object pose differences, and supports the operation of single or multiple objects.
    \item \textbf{Hinged Manipulations}: Focuses on tasks involving hinged items, such as opening doors or closing drawers.
    \item \textbf{Flexible Object Manipulations}: Targets tasks involving flexible materials, such as fabric, with typical examples including folding clothes or organizing textiles.
    \item \textbf{Fine Manipulations}: Tasks requiring high-precision control, such as pouring water, pressing buttons, or retrieving clothes from a clothesline.
    \item \textbf{Dual-arm Collaborative Manipulations}: Tasks requiring coordination between two robotic arms, such as transferring a pot from table to stove.
\end{enumerate}
Each task is composed of multiple fine-grained subtask segments, and due to the overlap and shared components among different task categories, the dataset presents both high diversity and significant complexity, making it a rich and challenging benchmark for robotic manipulation research. Figure~\ref{fig:data}(c) shows the count statistics of each task.

\subsection{Diversity of Objects}

Training a generalist policy model requires learning from datasets that encompass a wide spectrum of tasks and interactive objects, enabling generalization across diverse real-world scenarios. As shown in Figure~\ref{fig:data}(b), FastUMI-100K includes a large variety of interactive objects with differing shapes, sizes, and colors, distributed across various household environments, such as living rooms, studies, bathrooms, and kitchens. For example, the living room scene includes items like trash cans, clothes, irons, hangers, cosmetics, and cabinets; the study scene includes items like bookshelves, books, and backpacks; the bathroom scene includes items like washing machines and toilets; and the kitchen scene includes items like kitchen appliances, tableware, and sandwiches. The dataset incorporates a diverse set of object types, including rigid, flexible, and articulated objects, as well as items requiring precise manipulation, such as plugs and sockets. This rich and heterogeneous object set contributes to complex and varied interaction data, which in turn facilitates the development of policy models with improved generalization capabilities.
\begin{table*}[ht]
\centering
\vspace{-15pt} 
\caption{Success rates for diffusion policy in different tasks and platforms.}
\label{tab:spaced_task_details}
\begin{tabular}{ccccc}
\toprule
\textbf{Task} & \textbf{\makecell{Sub-Task Description}} & \textbf{\makecell{Manipulation\\Type}} & \textbf{Platform} & \textbf{\makecell{Successful Rate of DP}} \\
\midrule
\multirow{4}{*}{Make Sandwich} & \multirow{2}{*}{\makecell{Place the lettuce leaves on the plate}} & \multirow{2}{*}{Pick-Place \& Rotation} & Flexiv Rizon4 & 60.00\% \\
& & & Xarm6 & 53.33\% \\
\cmidrule(lr){2-5}
& \multirow{2}{*}{\makecell{Place the bread slices on the plate}} & \multirow{2}{*}{Pick-Place \& Rotation} & Flexiv Rizon4 & 60.00\% \\
& & & Xarm6 & 60.00\% \\
\cmidrule(lr){1-5}
\multirow{2}{*}{Place Tableware} & \multirow{2}{*}{\makecell{Place the fork into the chopstick holder}} & \multirow{2}{*}{Pick-Place \& Rotation} & Flexiv Rizon4 & 40.00\% \\
& & & Xarm6 & 33.33\% \\
\cmidrule(lr){1-5}
\multirow{2}{*}{Heat Food} & \multirow{2}{*}{\makecell{Open the microwave door}} & \multirow{2}{*}{Hinged} & Flexiv Rizon4 & 66.67\% \\
& & & Xarm6 & 66.67\% \\
\cmidrule(lr){1-5}
Pour Water & \makecell{Pour the water from the bottle into the cup} & Rotation & Xarm6 & 66.67\% \\
\cmidrule(lr){1-5}
Storage Shoes & \makecell{Place the shoe in the open cabinet} & Pick-Place & Xarm6 & 40.00\% \\
\cmidrule(lr){1-5}
Wash Clothes & \makecell{Place the clothes into the washing machine} & Pick-Place \& Rotation & Flexiv Rizon4 & 33.33\% \\
\bottomrule
\end{tabular}

\end{table*}
 
\begin{table}[htbp]
    \centering
    \vspace{-15pt}
    \caption{Comparison of success rates for different ACT variants across representative tasks.}
    \label{tab:task_success_comparison}
    \begin{adjustbox}{max width=\linewidth}
    \begin{tabular}{@{}l cc cc@{}}
    \toprule
    & \multicolumn{2}{c}{\textbf{Joint}} & \multicolumn{2}{c}{\textbf{TCP}} \\
    \cmidrule(lr){2-3} \cmidrule(lr){4-5}
    \textbf{Task} & \textbf{ACT} & \textbf{Smooth-ACT} & \makecell{\textbf{PoseACT}\\ \textbf{(Absolute)}} & \makecell{\textbf{PoseACT}\\ \textbf{(Relative)}} \\
    \midrule
    Pick Bear    & 20.00\% & 60.00\% & 80.00\% & 73.33\% \\
    Sweep Trash  &  6.67\% & 26.67\% & 53.33\% & 60.00\% \\
    \bottomrule
    \end{tabular}
    \end{adjustbox}
\vspace{-15pt} 
\end{table}
\subsection{Diversity of Task Horizon}

The variation in task durations reflects the underlying information complexity and difficulty associated with each trajectory. In the FastUMI-100K dataset, the original video data is recorded at 60 Hz and subsequently downsampled to 20 Hz to maintain temporal alignment with other modalities. The trajectory durations range from 6 to 25 seconds, corresponding to 120 to 500 frames. Compared with teleoperation-based data collection methods, the data acquired through our proposed collection scheme exhibits significantly higher information density. This is achieved by eliminating action delays and avoiding the redundant command transmission typically present in teleoperation systems. As a result, the trajectories are more temporally compact, leading to smoother and more efficient continuous manipulation sequences. For the same manipulation task, the time required for this scheme is only one-fifth of that using the teleoperation-based collection method. For example, the time spent on the clothes folding task in AgiBot~\cite{bu2025agibot} is 50 seconds, while FastUMI completes the same task in only 10 seconds. This characteristic significantly reduces the cost of data collection. By leveraging the efficiency and smooth execution enabled by the FastUMI data collection scheme, the long-horizon tasks in this dataset consist of multiple atomic skill segments. This design makes the tasks in the dataset more diverse and realistic.

\subsection{Flexibility and Human-like Motion Characteristics}
In our dataset, the operator interacts with the environment by directly manipulating the gripper, without the need for VR devices or complex teleoperation setups. This allows for intuitive, first-person interaction with objects, grounded in the human operator’s subjective perspective and natural motor skills, while also providing immediate tactile feedback from physical contact. As a result, the FastUMI-100K dataset differs fundamentally from those collected via traditional teleoperation, offering greater flexibility, higher fidelity of human-like motion, and more naturalistic interaction patterns. As shown in Figure~\ref{fig:data}(d) and (e), we compare the trajectory and angular variations per unit time between the FastUMI-100K and similar data collected from AgiBot~\cite{bu2025agibot}. This comparison result indicates that our data is closer to the motion characteristics of real humans when performing manipulation tasks than mainstream teleoperation data. This is crucial for training more human-like robots to perform fine manipulation in the future.

\section{Experimental Evaluation}
\begin{table*}[ht]
\centering
\vspace{-15pt}
\caption{Success rates for $\pi_0$-base finetuned in different tasks on Xarm6.}
\label{tab:pi0_task_details}
\begin{tabular}{cccccccccc}
\toprule
\textbf{Manipulation Type} & \multicolumn{3}{c}{\textbf{Hinged}} &  & \multicolumn{3}{c}{\textbf{Pick-Place}} &  & \multicolumn{1}{c}{\textbf{Hinged and Pick-Place}} \\
\cline{2-4} \cline{6-8} \cline{10-10}
\textbf{Task} & \makecell{\textbf{Open} \\ \textbf{Drawer}} & \makecell{\textbf{Open} \\ \textbf{Roaster}} & \makecell{\textbf{Open} \\ \textbf{Container}} &  & \makecell{\textbf{Rearrange} \\ \textbf{Coke}} & \makecell{\textbf{Unplug} \\ \textbf{Charger}} & \makecell{\textbf{Pick Series} \\ \textbf{(Bear/Lid/Cup)}} &  & \makecell{\textbf{Hotdog} \\ \textbf{in Roaster}} \\
\midrule
\textbf{Success Rate} & 80.00\% & 86.67\% & 73.33\% &  & 93.33\% & 93.33\% & \makecell{80.00 / 80.00 / 93.33\%} &  & 80.00\% \\
\bottomrule
\end{tabular}
\vspace{-10pt} 
\end{table*}

\begin{table}[ht]
\centering
\caption{Success rates for $\pi_0$ finetuned in different long-horizon tasks on Flexiv Rizon4.}
\label{tab:pi0_task_details_on_flexiv}
\setlength{\tabcolsep}{0.0pt} 
\begin{tabular}{@{}cccc@{}}
\toprule
\textbf{Task} & \textbf{Sub-Task Description} & \textbf{\makecell{Manipulation\\Type}} & \textbf{Successful Rate} \\
\midrule
\multirow{2}{*}{Make Sandwich} & Place lettuce on the plate & \multirow{2}{*}{~P-P \& Rotation} & 100.00\% \\
& Place bread on the plate & & 73.33\% \\
\midrule
\multirow{2}{*}{Wash Clothes} & Grasp clothes into washer & ~P-P \& Rotation & 93.33\% \\
& Close washer door & Hinged & 60.00\% \\
\midrule
\multirow{3}{*}{Heat Food} & Open microwave door & Hinged & 100.00\% \\
& Put bread into microwave & ~P-P \& Rotation & 0.00\% \\
& Close microwave door & Hinged & 0.00\% \\
\bottomrule
\end{tabular}
\vspace{-15pt} 
\end{table}

To systematically validate the effectiveness and versatility of the FastUMI-100K dataset, we conduct a series of comprehensive experiments leveraging diverse robotic manipulation learning methods. These experiments aim to assess the dataset’s ability to support policy training across single-task scenarios, cross-embodiment transfer, while also verifying the impact on Vision-Language-Action (VLA) models. Specifically, we evaluated the performance of single-task imitation learning models like Diffusion Policy (DP)~\cite{chi2023diffusion} and Action Chunking with Transformers (ACT)~\cite{zhao2023learning} in our dataset, explored the cross-platform applicability of the same UMI-style training data without fine-tuning using any specific robotic arm data, and validated the compatibility of FastUMI-100K with $\pi_0$ model~\cite{black2024pi0visionlanguageactionflowmodel} through fine-tuning experiments. Each task is repeated 15 times on different models, and the performance is quantitatively evaluated by calculating the success rate.

\subsection{Single-task Imitation Learning}
To evaluate the effectiveness of FastUMI-100K in single-task imitation learning, we select a diverse set of tasks and conduct a comprehensive policy evaluation using two representative single-task learning models, \emph{e.g.}, ACT and Diffusion-Policy. The task difficulties range from the simple basic and composite manipulation tasks to difficult hinged and fine-grained manipulation tasks. Due to the limitation of Diffusion Policy in handling long-horizon tasks, we choose to split long-horizon tasks into independent subtasks for experimental verification. This experiment demonstrates the validity of the UMI-style data in achieving high-quality training even under efficient data collection conditions. The experiments are primarily conducted on the Xarm6 and Flexiv Rizon4 platform.
\begin{figure}
\centering
\includegraphics[width=0.48\textwidth]{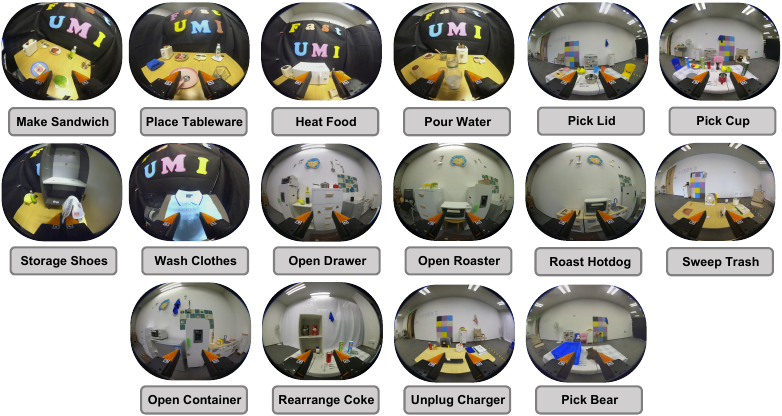}
\vskip -0.15in  
\caption{All 16 tasks evaluated in our experiment.}
\label{fig:exp_tasks}
\vskip -0.35in    
\end{figure}

As shown in Table~\ref{tab:spaced_task_details}, the success rates of the Diffusion Policy across seven different subtasks are presented, demonstrating that even with the most basic baseline model, our data remains reliable. Compared with previous datasets that focus on simple pick-and-place tasks under constrained scenarios, our dataset prioritizes realistic object manipulations, scene arrangements and sufficient diversity. The human-like nature of the motions in FastUMI-100K imposes greater demands on the inverse dynamics capabilities of robotic arms, as well as on their manipulation space. These characteristics collectively raise the complexity of the learning problem, requiring more advanced and robust algorithms. At the same time, this human-like nature aligns better with the future needs of robots in real-world scenarios and the evolving algorithmic requirements for such data.
Table~\ref{tab:task_success_comparison}, sourced from the pioneering work FastUMI~\cite{liu2024fastumi}, compares the training differences between joint and TCP (Tool Center Point) in the ACT algorithm. It shows that the TCP pose can robustly capture the shape and dynamic features of the trajectory, highlighting that our UMI-Style dataset, which is independent of the robot's body and utilizes end-effector states, allows for a more detailed representation of the dexterity and human-like motion characteristics of real-world manipulations.

\subsection{Cross-platform Deployment}
To validate the cross-platform transferability of the FastUMI-100K dataset and investigate the generalization performance of UMI-style data across different robotic embodiments, we design a cross robotic embodiment evaluation experiment. Specifically, models trained using single-task Diffusion Policy are transferred to two distinct robotic platforms (including  Xarm6 and  Flexiv Rizon4), both with different brands and structures, to perform the same tasks. During the transfer process, no additional fine-tuning is applied to the target robots, with only the coordinate system mapping of the end-effector being adjusted. 

As shown in Table~\ref{tab:spaced_task_details}, we conduct cross robotic embodiment experiments on four subtasks in task \texttt{Make Sandwich}, \texttt{Place Tableware}, and \texttt{Heat Food}. We found that since data collected by FastUMI is not constrained by the physical limitations of any specific robotic embodiment, human-collected data may exceed the manipulation range of robotic arms with smaller workspaces. To address this issue, when filtering data, a three-dimensional bounding box can be constructed based on the rated workspace of the target robot. If the main part of a trajectory exceeds the predefined bounding box, it is considered outside the physical reach of the robotic arm and should be filtered out. For example, the task ~\texttt{Wash Clothes} is completely beyond the workspace of the Xarm6 platform, thus preventing successful manipulation. This simple yet effective screening strategy enables the FastUMI-100K dataset to be flexibly deployed across robotic platforms with varying morphologies and workspace constraints.

\subsection{Fine-tuning on VLA Large Models}
To verify the compatibility effects of UMI-style data with existing robotic manipulation data in training the VLA model, we establishes the $\pi_0$-base pre-trained model as a baseline. The experimental design is presented where the FastUMI-100K dataset is used for fine-tuning a model initially pre-trained with traditional non-UMI data. The system evaluates the impact of joint training on model performance.

As shown in Table~\ref{tab:pi0_task_details}, we first selected nine short-horizon tasks from different manipulation types for experimentation. Each type of task achieved a high success rate, indicating that the fisheye perspective is equally capable of capturing the spatial state of objects and richer contextual information as compared to the commonly used RGB images in existing datasets. This aligns with the findings in FastUMI, where fisheye lenses installed on the end-effector and multi-view camera setups demonstrated comparable performance. It confirms that the UMI-style dataset can be jointly trained with conventional data in a VLA model without compromising model performance.

To demonstrate the effectiveness of long-horizon FastUMI-100K data in training VLA models, we conduct fine-tuning experiments on the $\pi_0$ base model using three complete long-horizon tasks, each consisting of multiple short-horizon subtasks. Unlike the Diffusion Policy experiments, we use a staged success rate continuous recording approach for testing a complete long-horizon task. As shown in Table~\ref{tab:pi0_task_details_on_flexiv}, for the tasks~\texttt{Make Sandwich} and ~\texttt{Wash Clothes}, the first stage shows high success rates, followed by a slight decline in accuracy during the second stage due to the accumulation of inference errors. Interestingly, for the ~\texttt{Heat Food} task, the success rate is high in the first stage. However, during the transition to the second stage, the model encounter a local failure between similar observations, such as pushing the microwave door open and grabbing the bread. As a result, the model is unable to proceed with the remaining actions. Since the dataset does not provide a fixed third-person perspective for global state information, the presence of similar visual observations across different stages of a trajectory can impact model inference, especially for models relying on first-person wrist perspectives. Future VLA models that incorporate richer historical and sequential information may be better equipped to overcome current limitations in generalization and long-horizon inference. As such models evolve, FastUMI-100K is expected to remain a valuable resource, continually contributing to improvements in model performance and robustness across diverse manipulation tasks.

\section{Conclusion and Future Work}
This paper presents the large-scale UMI-style dataset---FastUMI-100K, aimed at addressing the limitations of existing robotic manipulation datasets, such as insufficient scalability, poor trajectory smoothness, and limited adaptability across robotic embodiments. Collected using the FastUMI system, which features a modular, hardware-agnostic design and integrated lightweight tracking, the dataset encompasses over 100K demonstration trajectories across common household environments, covering 54 tasks and hundreds of objects. Experimental evaluations confirm FastUMI-100K’s effectiveness: it enables high policy success rates for single-task imitation learning (\emph{e.g.}, Diffusion Policy, ACT), supports cross-platform deployment across distinct robotic architectures with only coordinate mapping adjustments, and is compatible with VLA large models (\emph{e.g.}, $\pi_0$) for fine-tuning. These results validate the dataset’s robustness, adaptability, and real-world applicability, providing a valuable resource for advancing data-driven robotic manipulation.

In the future, we will leverage the FastUMI-100K dataset to further expand dual-arm collaborative manipulation and achieve faster, more dexterous, flexible, and human-like movements. Additionally, we will conduct further fusion training of UMI-style data and common data on VLA models to enable more complex long-horizon task reasoning.

\bibliographystyle{IEEEtran}
\bibliography{IEEEabrv,reference}

\begin{thebibliography}{10}
\providecommand{\url}[1]{#1}
\csname url@samestyle\endcsname
\providecommand{\newblock}{\relax}
\providecommand{\bibinfo}[2]{#2}
\providecommand{\BIBentrySTDinterwordspacing}{\spaceskip=0pt\relax}
\providecommand{\BIBentryALTinterwordstretchfactor}{4}
\providecommand{\BIBentryALTinterwordspacing}{\spaceskip=\fontdimen2\font plus
\BIBentryALTinterwordstretchfactor\fontdimen3\font minus \fontdimen4\font\relax}
\providecommand{\BIBforeignlanguage}[2]{{%
\expandafter\ifx\csname l@#1\endcsname\relax
\typeout{** WARNING: IEEEtran.bst: No hyphenation pattern has been}%
\typeout{** loaded for the language `#1'. Using the pattern for}%
\typeout{** the default language instead.}%
\else
\language=\csname l@#1\endcsname
\fi
#2}}
\providecommand{\BIBdecl}{\relax}
\BIBdecl

\bibitem{black2024pi0visionlanguageactionflowmodel}
\BIBentryALTinterwordspacing
K.~Black, N.~Brown, D.~Driess, A.~Esmail, M.~Equi, C.~Finn, N.~Fusai, L.~Groom, K.~Hausman, B.~Ichter, S.~Jakubczak, T.~Jones, L.~Ke, S.~Levine, A.~Li-Bell, M.~Mothukuri, S.~Nair, K.~Pertsch, L.~X. Shi, J.~Tanner, Q.~Vuong, A.~Walling, H.~Wang, and U.~Zhilinsky, ``$\pi_0$: A vision-language-action flow model for general robot control,'' 2024. [Online]. Available: \url{https://arxiv.org/abs/2410.24164}
\BIBentrySTDinterwordspacing

\bibitem{qu2025embodiedonevision}
D.~Qu, H.~Song, Q.~Chen, Z.~Chen, X.~Gao, X.~Ye, Q.~Lv, M.~Shi, G.~Ren, C.~Ruan \emph{et~al.}, ``Embodiedonevision: Interleaved vision-text-action pretraining for general robot control,'' \emph{arXiv preprint arXiv:2508.21112}, 2025.

\bibitem{qu2025spatialvla}
D.~Qu, H.~Song, Q.~Chen, Y.~Yao, X.~Ye, Y.~Ding, Z.~Wang, J.~Gu, B.~Zhao, D.~Wang \emph{et~al.}, ``Spatialvla: Exploring spatial representations for visual-language-action model,'' \emph{arXiv preprint arXiv:2501.15830}, 2025.

\bibitem{team2024octo}
O.~M. Team, D.~Ghosh, H.~Walke, K.~Pertsch, K.~Black, O.~Mees, S.~Dasari, J.~Hejna, T.~Kreiman, C.~Xu \emph{et~al.}, ``Octo: An open-source generalist robot policy,'' \emph{arXiv preprint arXiv:2405.12213}, 2024.

\bibitem{liu2024rdt}
S.~Liu, L.~Wu, B.~Li, H.~Tan, H.~Chen, Z.~Wang, K.~Xu, H.~Su, and J.~Zhu, ``Rdt-1b: a diffusion foundation model for bimanual manipulation,'' \emph{arXiv preprint arXiv:2410.07864}, 2024.

\bibitem{kim2024openvla}
M.~J. Kim, K.~Pertsch, S.~Karamcheti, T.~Xiao, A.~Balakrishna, S.~Nair, R.~Rafailov, E.~Foster, G.~Lam, P.~Sanketi \emph{et~al.}, ``Openvla: An open-source vision-language-action model,'' \emph{arXiv preprint arXiv:2406.09246}, 2024.

\bibitem{song2025hume}
H.~Song, D.~Qu, Y.~Yao, Q.~Chen, Q.~Lv, Y.~Tang, M.~Shi, G.~Ren, M.~Yao, B.~Zhao \emph{et~al.}, ``Hume: Introducing system-2 thinking in visual-language-action model,'' \emph{arXiv preprint arXiv:2505.21432}, 2025.

\bibitem{walke2023bridgedata}
H.~R. Walke, K.~Black, T.~Z. Zhao, Q.~Vuong, C.~Zheng, P.~Hansen-Estruch, A.~W. He, V.~Myers, M.~J. Kim, M.~Du \emph{et~al.}, ``Bridgedata v2: A dataset for robot learning at scale,'' in \emph{Conference on Robot Learning}.\hskip 1em plus 0.5em minus 0.4em\relax PMLR, 2023, pp. 1723--1736.

\bibitem{o2024open}
A.~O’Neill, A.~Rehman, A.~Maddukuri, A.~Gupta, A.~Padalkar, A.~Lee, A.~Pooley, A.~Gupta, A.~Mandlekar, A.~Jain \emph{et~al.}, ``Open x-embodiment: Robotic learning datasets and rt-x models: Open x-embodiment collaboration 0,'' in \emph{2024 IEEE International Conference on Robotics and Automation (ICRA)}.\hskip 1em plus 0.5em minus 0.4em\relax IEEE, 2024, pp. 6892--6903.

\bibitem{wu2024robomind}
K.~Wu, C.~Hou, J.~Liu, Z.~Che, X.~Ju, Z.~Yang, M.~Li, Y.~Zhao, Z.~Xu, G.~Yang \emph{et~al.}, ``Robomind: Benchmark on multi-embodiment intelligence normative data for robot manipulation,'' \emph{arXiv preprint arXiv:2412.13877}, 2024.

\bibitem{bu2025agibot}
Q.~Bu, J.~Cai, L.~Chen, X.~Cui, Y.~Ding, S.~Feng, S.~Gao, X.~He, X.~Hu, X.~Huang \emph{et~al.}, ``Agibot world colosseo: A large-scale manipulation platform for scalable and intelligent embodied systems,'' \emph{arXiv preprint arXiv:2503.06669}, 2025.

\bibitem{fang2023rh20t}
H.-S. Fang, H.~Fang, Z.~Tang, J.~Liu, C.~Wang, J.~Wang, H.~Zhu, and C.~Lu, ``Rh20t: A comprehensive robotic dataset for learning diverse skills in one-shot,'' \emph{arXiv preprint arXiv:2307.00595}, 2023.

\bibitem{ebert2021bridge}
F.~Ebert, Y.~Yang, K.~Schmeckpeper, B.~Bucher, G.~Georgakis, K.~Daniilidis, C.~Finn, and S.~Levine, ``Bridge data: Boosting generalization of robotic skills with cross-domain datasets,'' \emph{arXiv preprint arXiv:2109.13396}, 2021.

\bibitem{brohan2022rt}
A.~Brohan, N.~Brown, J.~Carbajal, Y.~Chebotar, J.~Dabis, C.~Finn, K.~Gopalakrishnan, K.~Hausman, A.~Herzog, J.~Hsu \emph{et~al.}, ``Rt-1: Robotics transformer for real-world control at scale,'' \emph{arXiv preprint arXiv:2212.06817}, 2022.

\bibitem{wu2024gello}
P.~Wu, Y.~Shentu, Z.~Yi, X.~Lin, and P.~Abbeel, ``Gello: A general, low-cost, and intuitive teleoperation framework for robot manipulators,'' in \emph{2024 IEEE/RSJ International Conference on Intelligent Robots and Systems (IROS)}.\hskip 1em plus 0.5em minus 0.4em\relax IEEE, 2024, pp. 12\,156--12\,163.

\bibitem{iyer2024open}
A.~Iyer, Z.~Peng, Y.~Dai, I.~Guzey, S.~Haldar, S.~Chintala, and L.~Pinto, ``Open teach: A versatile teleoperation system for robotic manipulation,'' \emph{arXiv preprint arXiv:2403.07870}, 2024.

\bibitem{morgan2021vision}
A.~S. Morgan, B.~Wen, J.~Liang, A.~Boularias, A.~M. Dollar, and K.~Bekris, ``Vision-driven compliant manipulation for reliable, high-precision assembly tasks,'' \emph{arXiv preprint arXiv:2106.14070}, 2021.

\bibitem{wen2022you}
B.~Wen, W.~Lian, K.~Bekris, and S.~Schaal, ``You only demonstrate once: Category-level manipulation from single visual demonstration,'' \emph{arXiv preprint arXiv:2201.12716}, 2022.

\bibitem{song2020grasping}
S.~Song, A.~Zeng, J.~Lee, and T.~Funkhouser, ``Grasping in the wild: Learning 6dof closed-loop grasping from low-cost demonstrations,'' \emph{IEEE Robotics and Automation Letters}, vol.~5, no.~3, pp. 4978--4985, 2020.

\bibitem{chi2024universal}
C.~Chi, Z.~Xu, C.~Pan, E.~Cousineau, B.~Burchfiel, S.~Feng, R.~Tedrake, and S.~Song, ``Universal manipulation interface: In-the-wild robot teaching without in-the-wild robots,'' \emph{arXiv preprint arXiv:2402.10329}, 2024.

\bibitem{liu2024fastumi}
K.~Liu, C.~Guan, Z.~Jia, Z.~Wu, X.~Liu, T.~Wang, S.~Liang, P.~Chen, P.~Zhang, H.~Song \emph{et~al.}, ``Fastumi: A scalable and hardware-independent universal manipulation interface with dataset,'' \emph{arXiv preprint arXiv:2409.19499}, 2024.

\bibitem{liu2025vitamin}
F.~Liu, C.~Li, Y.~Qin, A.~Shaw, J.~Xu, P.~Abbeel, and R.~Chen, ``Vitamin: Learning contact-rich tasks through robot-free visuo-tactile manipulation interface,'' \emph{arXiv preprint arXiv:2504.06156}, 2025.

\bibitem{wu2025freetacman}
L.~Wu, C.~Yu, J.~Ren, L.~Chen, R.~Huang, G.~Gu, and H.~Li, ``Freetacman: Robot-free visuo-tactile data collection system for contact-rich manipulation,'' \emph{arXiv preprint arXiv:2506.01941}, 2025.

\bibitem{huang20243d}
B.~Huang, Y.~Wang, X.~Yang, Y.~Luo, and Y.~Li, ``3d-vitac: Learning fine-grained manipulation with visuo-tactile sensing,'' \emph{arXiv preprint arXiv:2410.24091}, 2024.

\bibitem{aldaco2024aloha}
J.~Aldaco, T.~Armstrong, R.~Baruch, J.~Bingham, S.~Chan, K.~Draper, D.~Dwibedi, C.~Finn, P.~Florence, S.~Goodrich \emph{et~al.}, ``Aloha 2: An enhanced low-cost hardware for bimanual teleoperation,'' \emph{arXiv preprint arXiv:2405.02292}, 2024.

\bibitem{ding2024bunny}
R.~Ding, Y.~Qin, J.~Zhu, C.~Jia, S.~Yang, R.~Yang, X.~Qi, and X.~Wang, ``Bunny-visionpro: Real-time bimanual dexterous teleoperation for imitation learning,'' \emph{arXiv preprint arXiv:2407.03162}, 2024.

\bibitem{dasari2019robonet}
S.~Dasari, F.~Ebert, S.~Tian, S.~Nair, B.~Bucher, K.~Schmeckpeper, S.~Singh, S.~Levine, and C.~Finn, ``Robonet: Large-scale multi-robot learning,'' \emph{arXiv preprint arXiv:1910.11215}, 2019.

\bibitem{khazatsky2024droid}
A.~Khazatsky, K.~Pertsch, S.~Nair, A.~Balakrishna, S.~Dasari, S.~Karamcheti, S.~Nasiriany, M.~K. Srirama, L.~Y. Chen, K.~Ellis \emph{et~al.}, ``Droid: A large-scale in-the-wild robot manipulation dataset,'' \emph{arXiv preprint arXiv:2403.12945}, 2024.

\bibitem{zitkovich2023rt}
B.~Zitkovich, T.~Yu, S.~Xu, P.~Xu, T.~Xiao, F.~Xia, J.~Wu, P.~Wohlhart, S.~Welker, A.~Wahid \emph{et~al.}, ``Rt-2: Vision-language-action models transfer web knowledge to robotic control,'' in \emph{Conference on Robot Learning}.\hskip 1em plus 0.5em minus 0.4em\relax PMLR, 2023, pp. 2165--2183.

\bibitem{belkhale2023data}
S.~Belkhale, Y.~Cui, and D.~Sadigh, ``Data quality in imitation learning,'' \emph{Advances in neural information processing systems}, vol.~36, pp. 80\,375--80\,395, 2023.

\bibitem{belkhale2024rt}
S.~Belkhale, T.~Ding, T.~Xiao, P.~Sermanet, Q.~Vuong, J.~Tompson, Y.~Chebotar, D.~Dwibedi, and D.~Sadigh, ``Rt-h: Action hierarchies using language,'' \emph{arXiv preprint arXiv:2403.01823}, 2024.

\bibitem{chi2023diffusion}
C.~Chi, Z.~Xu, S.~Feng, E.~Cousineau, Y.~Du, B.~Burchfiel, R.~Tedrake, and S.~Song, ``Diffusion policy: Visuomotor policy learning via action diffusion,'' \emph{The International Journal of Robotics Research}, p. 02783649241273668, 2023.

\bibitem{zhao2023learning}
T.~Z. Zhao, V.~Kumar, S.~Levine, and C.~Finn, ``Learning fine-grained bimanual manipulation with low-cost hardware,'' \emph{arXiv preprint arXiv:2304.13705}, 2023.

\end{thebibliography}

\end{document}